%% file: main.tex
\documentclass{article}

\usepackage{times}
\usepackage{graphicx} 
\usepackage{subfigure} 

\usepackage{natbib}

\usepackage{algorithm}
\usepackage{algorithmic}

\usepackage{hyperref}

\usepackage{amsfonts}
\usepackage{amsmath,mathtools}
\usepackage{multirow}
\usepackage{booktabs}
\usepackage[normalem]{ulem}
\usepackage{amssymb}
\usepackage{parskip}
\newcommand{\E}{\mathbb{E}}
\usepackage{epsfig}

\usepackage[accepted]{icml2017}

\icmltitlerunning{Confident Multiple Choice Learning}

\begin{document} 

\twocolumn[
\icmltitle{Confident Multiple Choice Learning}

\icmlsetsymbol{equal}{*}

\begin{icmlauthorlist}
\icmlauthor{Kimin Lee}{to}
\icmlauthor{Changho Hwang}{to}
\icmlauthor{KyoungSoo Park}{to}
\icmlauthor{Jinwoo Shin}{to}
\end{icmlauthorlist}

\icmlaffiliation{to}{School of Electrical Engineering, Korea Advanced Institute of Science and Technology (KAIST), Daejeon, Repulic of Korea}

\icmlcorrespondingauthor{{Jinwoo Shin}}{jinwoos@kaist.ac.kr}
\icmlkeywords{boring formatting information, machine learning, ICML}

\vskip 0.3in
]



\printAffiliationsAndNotice{}  

\input{abstract}
\input{intro}
\input{uncertainty}
\input{model}

\input{feature-sharing}
\input{experiments}

\input{conclusion}
\input{Ack}
\bibliography{ref}
\bibliographystyle{icml2017}
\input{appendix}

\end{document}

%% file: abstract.tex
\begin{abstract}
Ensemble methods are arguably the most trustworthy 
techniques for boosting the performance of machine learning models.
Popular independent ensembles ({IE}) relying on na{\"i}ve averaging/voting scheme
have been 
of typical choice for most applications involving deep neural networks, but
they do not consider advanced collaboration among ensemble models.
In this paper,
we propose new ensemble methods
specialized for deep neural networks, 
called confident multiple choice learning ({CMCL}):
it is
a variant of multiple choice learning ({MCL}) via
addressing its overconfidence issue. 
In particular, the proposed major components of CMCL beyond the original MCL scheme
are 
(i) new loss, i.e., confident oracle loss, 
(ii) new architecture, i.e., feature sharing and 
(iii) new training method, i.e., stochastic labeling.
We demonstrate the effect of CMCL via experiments on the image classification on CIFAR and SVHN,
and the foreground-background segmentation on the iCoseg. 
In particular, CMCL using 5 residual networks provides $14.05\%$ and $6.60\%$ relative reductions in the top-1 error rates from the corresponding IE scheme for the classification task on CIFAR and SVHN, respectively.
\end{abstract}

%% file: intro.tex
\section{Introduction}
\label{sec:intro}

Ensemble methods have played a critical role in the machine learning community to obtain better predictive performance than what could be obtained from any of the constituent learning models alone,
e.g., Bayesian model/parameter averaging \cite{domingos2000bayesian}, boosting \cite{freund1999short} and bagging \cite{breiman1996bagging}. 
Recently, they have been successfully applied to enhancing
the power of many deep neural networks, e.g., 
80\% of top-5 best-performing teams on ILSVRC challenge 2016 \cite{imagenet12} employ ensemble methods.
They are easy and trustworthy to apply for most scenarios.
While there exists a long history on ensemble methods, 
the progress on developing more advanced ensembles specialized for deep neural networks has been slow. 
Despite continued efforts that apply various ensemble methods such as bagging and boosting to deep models, 
it has been observed that traditional independent ensembles (IE) which train models independently
with random initialization achieve the best performance \cite{ciregan2012multi,lee2015m}.
In this paper, we focus on developing more advanced ensembles for deep models utilizing the concept of multiple choice learning (MCL).

The MCL concept  was originally proposed in \cite{guzman2012multiple} under the scenario
when inference procedures are cascaded:
\begin{itemize}
    \item[(a)] First, generate a set of plausible outputs.
    \item[(b)] Then, pick the correct solution form the set.
\end{itemize}
For example, 
\citep{park2011n,batra2012diverse}
proposed human-pose estimation methods which produce multiple predictions and then refine them by employing a temporal model, and \citep{collins2005discriminative} proposed a sentence parsing method which re-ranks the output of an initial system which produces a set of plausible outputs \cite{huang2005better}.
In such scenarios, the goal of the first stage (a) is generating a set of plausible outputs such that at least one of them is correct for the second stage (b), e.g., human operators.
Under this motivation, MCL has been studied \cite{mclaistat,guzman2012multiple,lee2016stochastic}, 
where various applications have been demonstrated, e.g., image classification \cite{cifar09}, semantic segmentation \cite{everingham2010pascal} and image captioning \cite{lin2014microsoft}.
It trains an ensemble of multiple models by
minimizing the so-called {\em oracle loss}, only focusing on the most accurate prediction produced by them.
Consequently, it makes each model specialized for a certain subset
of data, not for the entire one similarly
as mixture-of-expert schemes \cite{jacobs1991adaptive}.

Although MCL focuses on the first stage (a) in cascaded scenarios and thus can produce diverse/plausible outputs, 
it might be not useful if one does not have a good scheme for the second stage (b).
One can use a certain average/voting scheme of the predictions made by models for (b),
but MCL using deep neural networks often fails to make a correct decision since each network tends to be overconfident in its prediction. Namely, the oracle error/loss of MCL is low, but its {\em top-1} error rate might be very high.

\noindent {\bf Contribution.}
To address the issue, we develop the concept of confident MCL (CMCL)
that does not lose any benefit of the original MCL, while its target loss and architecture are redesigned for making the second stage (b) easier. Specifically, it
targets to generate a set of diverse/plausible confident predictions from which one can pick the correct one using a simple average/voting scheme.
To this end, we first propose a new loss function, called {\em confident oracle loss}, for relaxing the overconfidence issue of MCL. Our key idea is to additionally 
minimize the Kullback-Leibler divergence from a predictive distribution to the uniform one in order to give confidence to non-specialized models.
Then, CMCL that minimizes the new loss can be efficiently trained like the original MCL for certain classes of models including neural networks,
via stochastic alternating minimization \cite{lee2016stochastic}.
Furthermore, when CMCL is applied to deep models,
we propose two additional {regularization} 
techniques for boosting its performance: {\em feature sharing}
and {\em stochastic labeling}.
Despite the new components,
we note that the training complexity of CMCL is 
almost same to that of MCL or IE.

We apply the new ensemble model trained by the new training scheme 
for several convolutional neural networks (CNNs) including
VGGNet \cite{vggnet}, GoogLeNet \cite{googlenet}, and ResNet \cite{resnet} for image classification on the CIFAR \cite{cifar09} and SVHN \cite{11SVHN} datasets, and fully-convolutional neural networks (FCNs) \cite{long2015fully} for foreground-background segmentation on the iCoseg dataset \cite{icoseg10}.
{First, for the image classification task,
CMCL outperforms all baselines, i.e., the traditional IE and the original MCL, in top-1 error rates.
In particular, CMCL of 5 ResNet with 20 layers provides $14.05\%$ and $6.60\%$ relative reductions in the top-1 error rates from the corresponding IE on CIFAR-10 and SVHN, respectively.}
Second, for the foreground-background segmentation task, CMCL using multiple FCNs with 4 layers also outperforms all baselines in top-1 error rates.
Each model trained by CMCL generates high-quality solutions by specializing for specific images while each model trained by IE does not.
We believe that our new approach should be of broader interest for many deep learning tasks requiring high accuracy.

{\bf Organization.}
In Section \ref{sec:relatedwork}, we introduce necessary backgrounds for multiple choice learning
and the corresponding loss function.
We describe the proposed loss and the corresponding training scheme in Section \ref{sec:model}.
Section \ref{sec:feature-gen} provides additional techniques for the proposed ensemble model.
Experimental results are reported in Section \ref{sec:exp}.

%% file: uncertainty.tex
\section{Preliminaries}
\label{sec:relatedwork}

\begin{figure*} [t] \centering
\vspace{-0.05in}
\subfigure[Multiple choice learning (MCL)]
{\epsfig{file=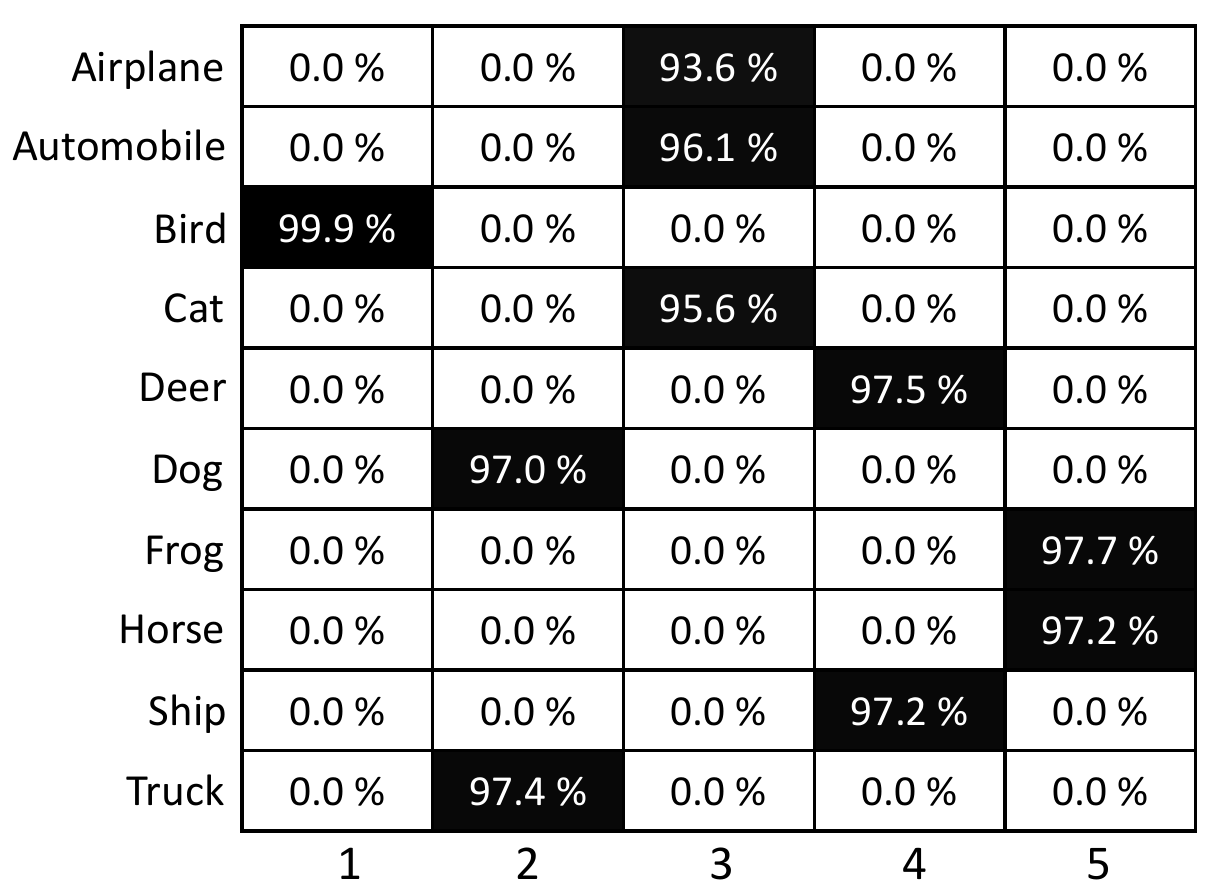, width=0.355\textwidth}\label{fig1:b}}
\,
\subfigure[Confident MCL (CMCL)]
{\epsfig{file=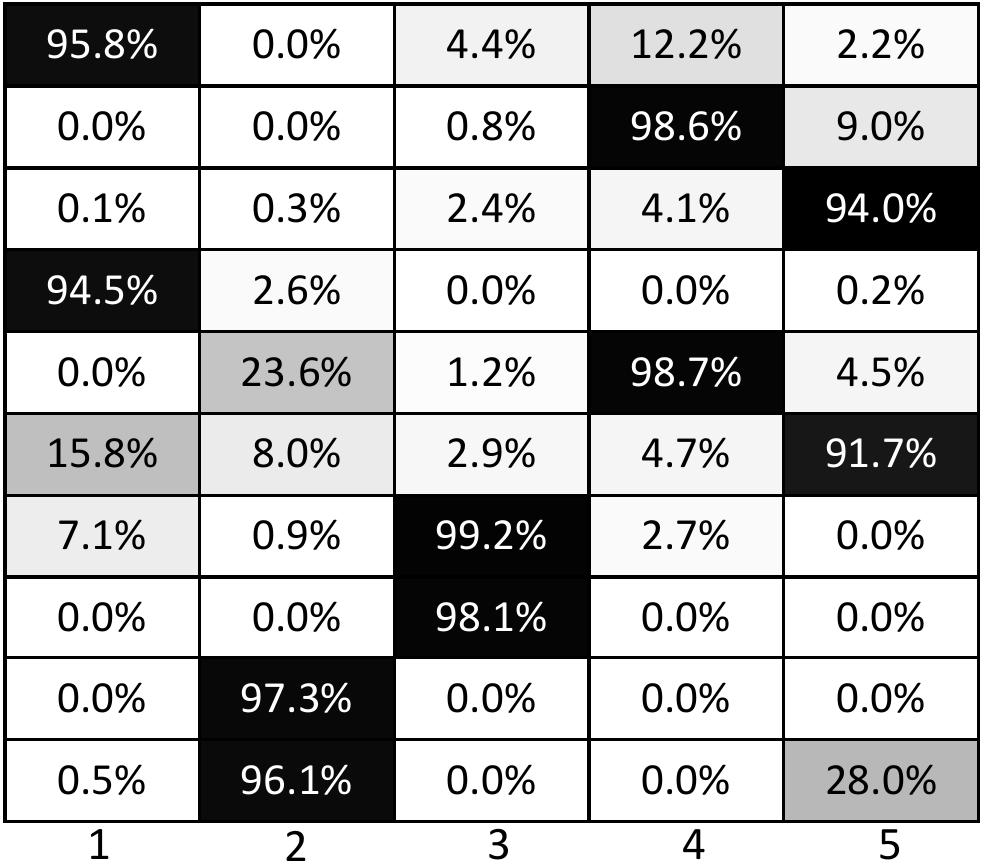, width=0.29\textwidth}\label{fig1:c}}
\,
\subfigure[Independent ensemble (IE)]
{\epsfig{file=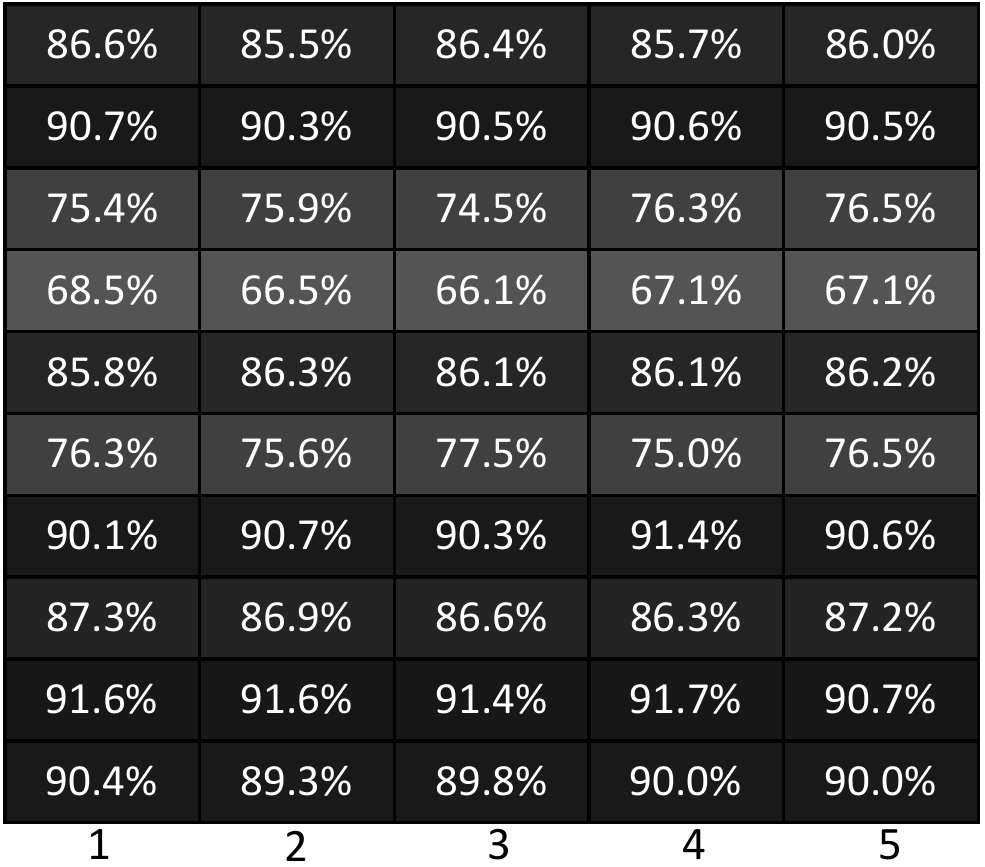, width=0.29\textwidth}\label{fig1:a}}
\caption{Class-wise test set accuracy of each ensemble model trained by various ensemble methods.
One can observe that most models trained by MCL and CMCL become specialists for certain classes while they are generalized in case of traditional IE.
}\label{fig1}
\vspace{-0.1in}
\end{figure*}

\subsection{Multiple Choice Learning}

In this section, we describe the basic concept of multiple choice learning (MCL) \cite{mclaistat,guzman2012multiple}.
Throughout this paper, we denote the set $\{1,\ldots, n \}$ by $[n]$ for positive integer $n$.
The MCL scheme is a type of ensemble learning that produces diverse outputs of high quality.
Formally, given a training dataset $\mathcal{D}=\{ (\mathbf{x}_i, y_i) \mid i \in [N],~  \mathbf{x}_i \in \mathcal{X},~ y_i \in \mathcal{Y} \}$,
we consider an ensemble of $M$ models $f$, i.e., $\left( f_1, \ldots, f_M \right)$.
For some task-specific loss function $\ell \left(y, f\left( \mathbf{x} \right) \right)$, the oracle loss over the dataset $D$ is defined as follows:
\begin{align}
L_{O} \left( \mathcal{D} \right) = & \sum \limits_{i=1}^N  \min \limits_{m\in [M]} \ell \left( y_i, f_{m} \left( \mathbf{x}_i \right) \right),
\label{eq:oracleloss}
\end{align}
while the traditional independent ensemble (IE) loss is 
\begin{align}
L_{E} \left( \mathcal{D} \right) = & \sum \limits_{i=1}^N  \sum\limits_{m\in [M]} \ell \left( y_i, f_{m} \left( \mathbf{x}_i \right) \right).
\label{eq:ensemloss}
\end{align}
If all models have the same capacity and one can obtain the (global) optimum of the IE loss with respect to the model parameters,
then all trained models should produce the same outputs, i.e., $f_1= \ldots = f_M$.
On the other hand, 
the oracle loss makes the most accurate model optimize the loss function $\ell \left(y, f\left( \mathbf{x} \right) \right)$ for each data $\mathbf{x}$.
Therefore, MCL produces diverse outputs of high quality by forcing each model to be specialized on a part of the entire dataset.

Minimizing the oracle loss \eqref{eq:oracleloss} is harder than minimizing the independent ensemble loss \eqref{eq:ensemloss}
since 
the $\min$ function is a non-continuous function.
To address the issue,
\citep{guzman2012multiple} proposed an iterative block coordinate decent algorithm and \citep{dey2015predicting} reformulated this problem as a submodular optimization task in which ensemble models are trained sequentially in a boosting-like manner.
However, when one considers an ensemble of deep neural networks,
it is challenging to apply these methods since they require either costly retraining or sequential training.
Recently, \citep{lee2016stochastic} overcame this issue by proposing a stochastic gradient descent (SGD) based algorithm. Throughout this paper, 
we primarily focus on 
ensembles of deep neural networks
and use the SGD algorithm for optimizing the oracle loss \eqref{eq:oracleloss} or its variants.

\subsection{Oracle Loss for Top-1 Choice}
\label{ssec:oracleloss}

The oracle loss \eqref{eq:oracleloss} used for MCL is useful for producing diverse/plausible outputs, but
it is often inappropriate for applications requiring a single choice, i.e., top-1 error.
This is because
ensembles of deep neural networks tend to be overconfident in their predictions, and
it is hard to judge a better solution from their outputs. 
To explain this in more detail, 
we evaluate the performance of ensembles of convolutional neural networks (CNNs)
for the image classification task on the CIFAR-10 dataset \cite{cifar09}.
We train ensembles of 5 CNNs (two convolutional layers followed by a fully-connected layer) using MCL.
We also 
train the models using traditional IE which trains each model independently under different random initializations.
Figure \ref{fig1} summarizes the class-wise test set accuracy of each ensemble member.
In the case of MCL, most models become specialists for certain classes (see Figure \ref{fig1:b}),
while they are generalized in the case of traditional IE as shown in Figure \ref{fig1:a}.
However, as expected, 
each model trained by MCL significantly outperforms for its specialized classes than that trained by IE.
For choosing a single output, similar to \citep{dropc,ciregan2012multi},
one can average the output probabilities from ensemble members trained by MCL,
but the corresponding top-1 classification error rate is often
very high (e.g., see Table \ref{tbl:comparison} in Section \ref{sec:exp}).
This is because each model trained by MCL is overconfident for its non-specialized classes.
To quantify this,
we also compute the entropy of the predictive distribution on the test data and use this to evaluate the quality of confidence/uncertainty level. 
Figure \ref{fig2:a} reports the entropy extracted from the predictive distribution of one of ensemble models trained by MCL.
One can observe that it has low entropy as expected for its specialized classes 
(i.e., classes that the model has a test accuracy higher than 90\%).
However, even for non-specialized classes, it also has low entropy.
Due to this, with respect to top-1 error rates, simple averaging of models trained by MCL performs much worse than that of IE. 
Such issue typically occurs in deep neural networks
since it is well known that they are poor at quantifying predictive uncertainties, 
and tend to be easily overconfident \cite{highconfidence15}.

%% file: model.tex
\section{Confident Multiple Choice Learning}
\label{sec:model}

\begin{figure*} [t] \centering
\vspace{-0.05in}
\subfigure[MCL]
{\epsfig{file=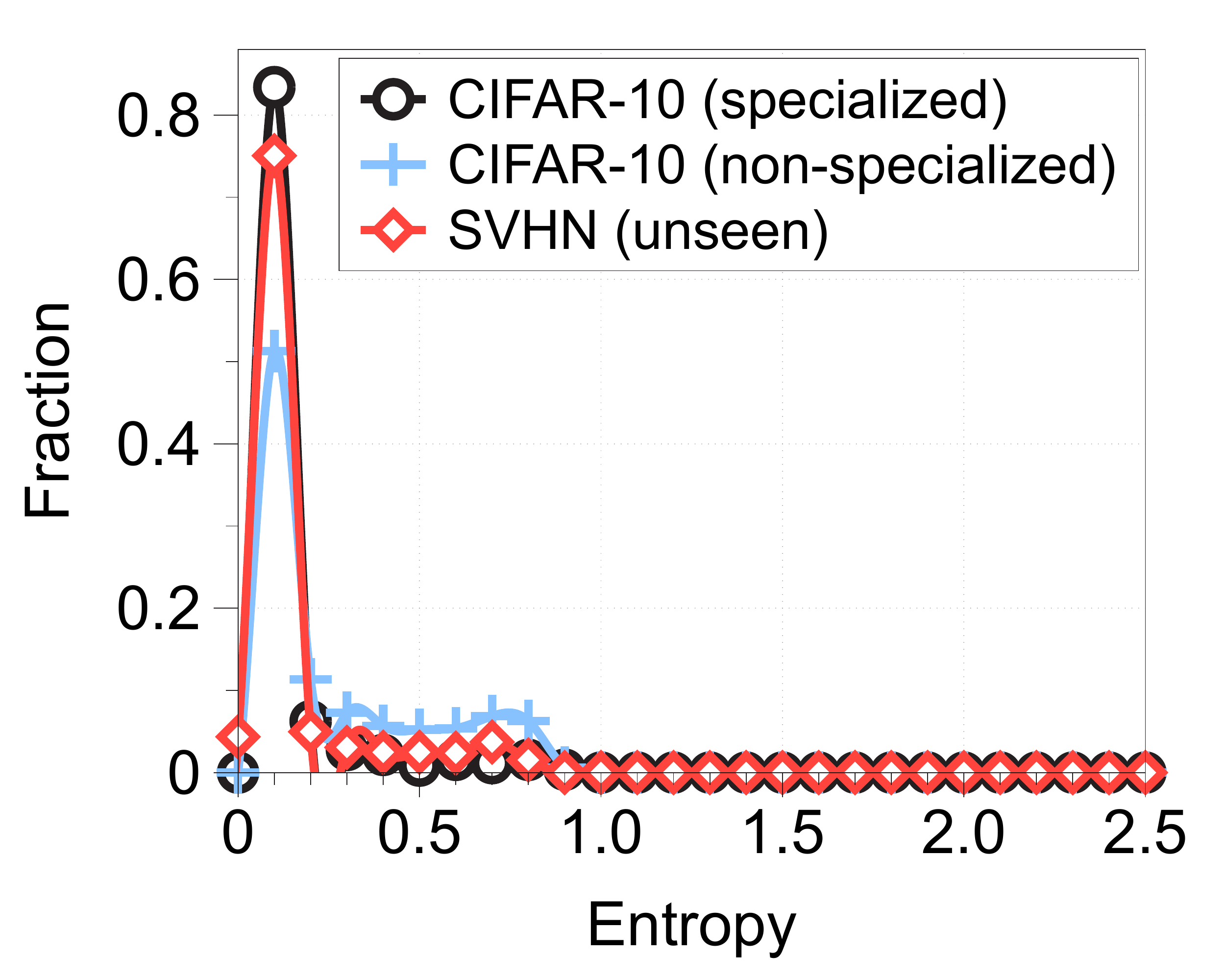, width=0.23\textwidth}\label{fig2:a}}
\,
\subfigure[CMCL]
{\epsfig{file=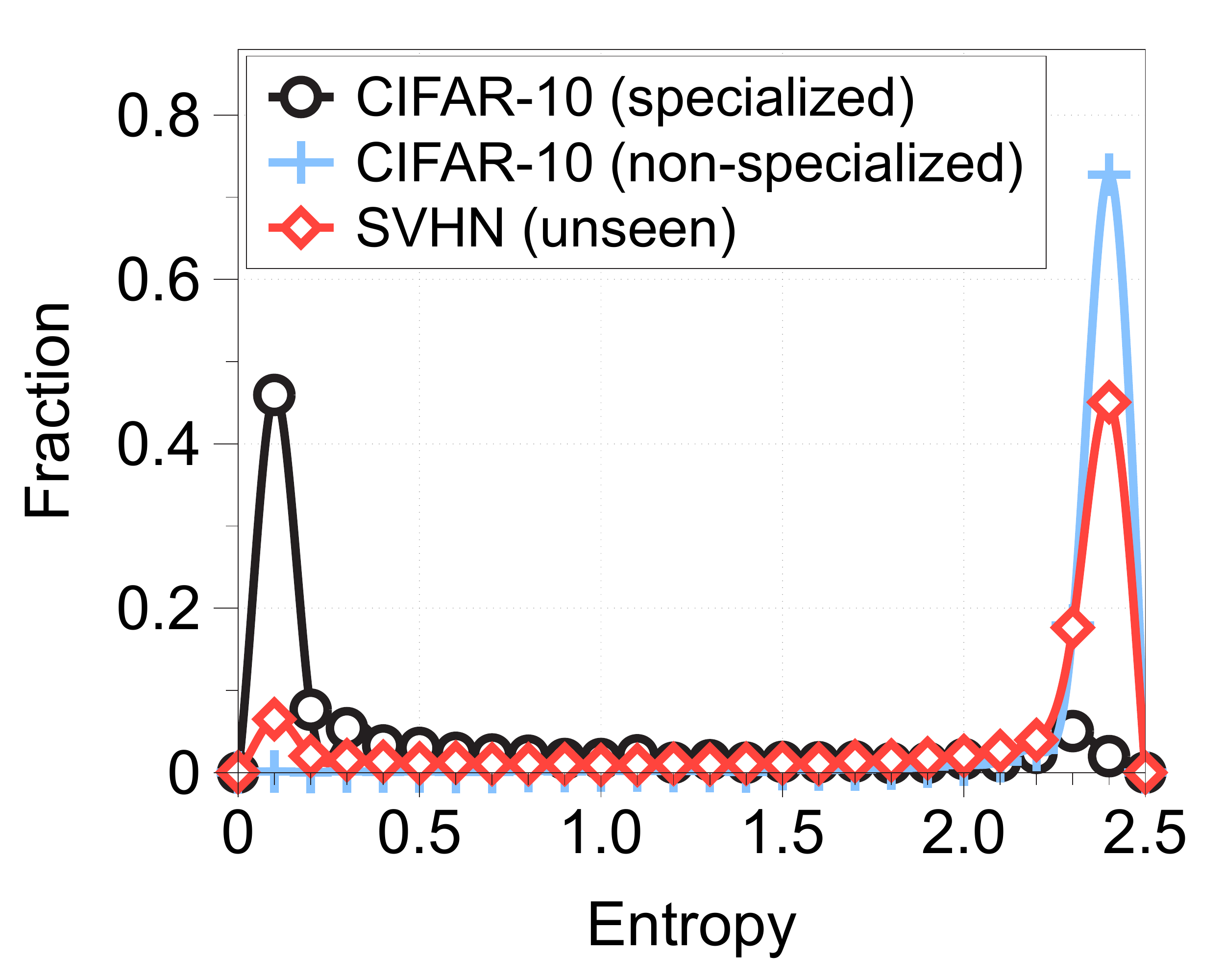, width=0.23\textwidth}\label{fig2:b}}
\,
\subfigure[IE with AT]
{\epsfig{file=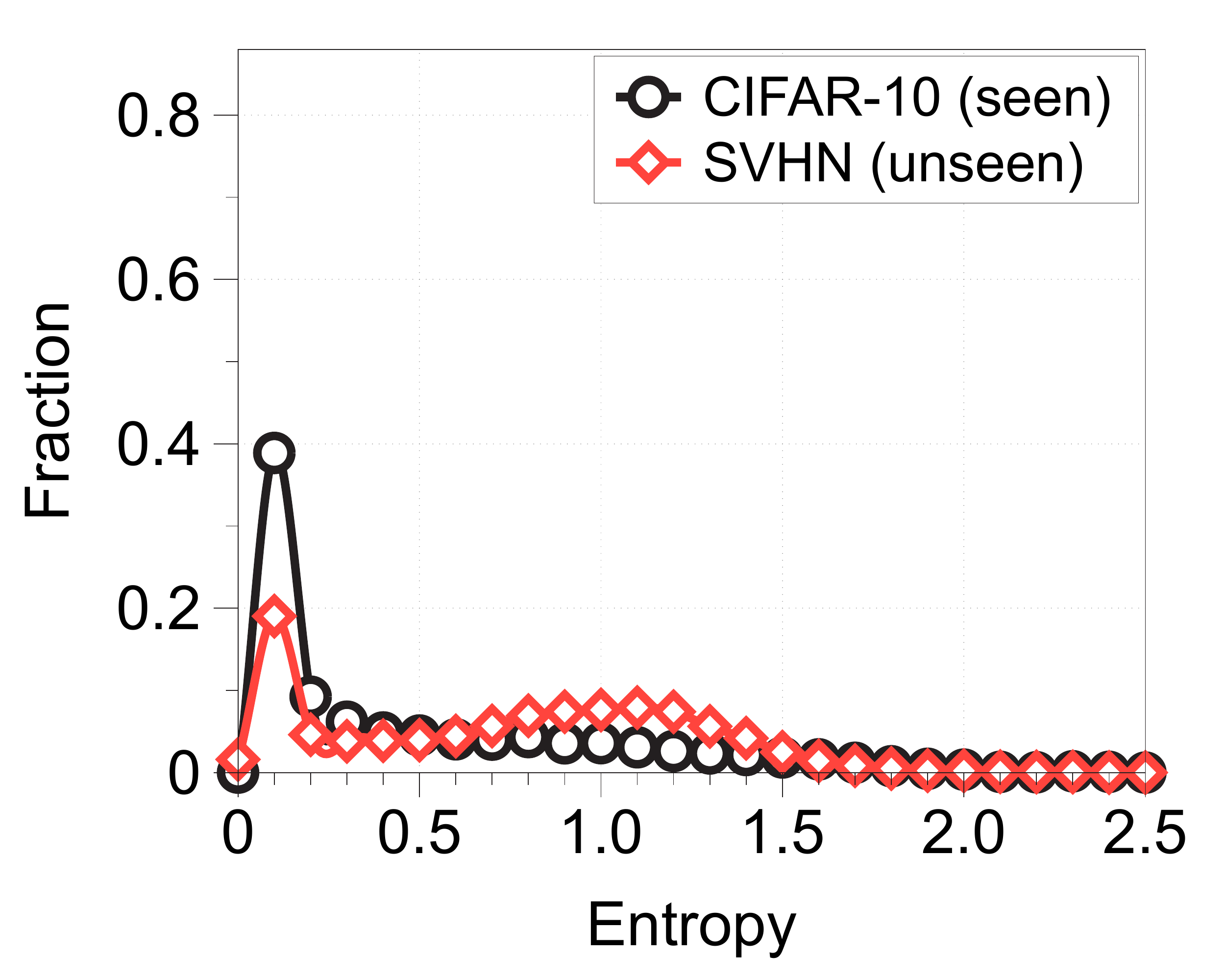, width=0.23\textwidth}\label{fig2:d}}
\,
\subfigure[Feature sharing]
{\epsfig{file=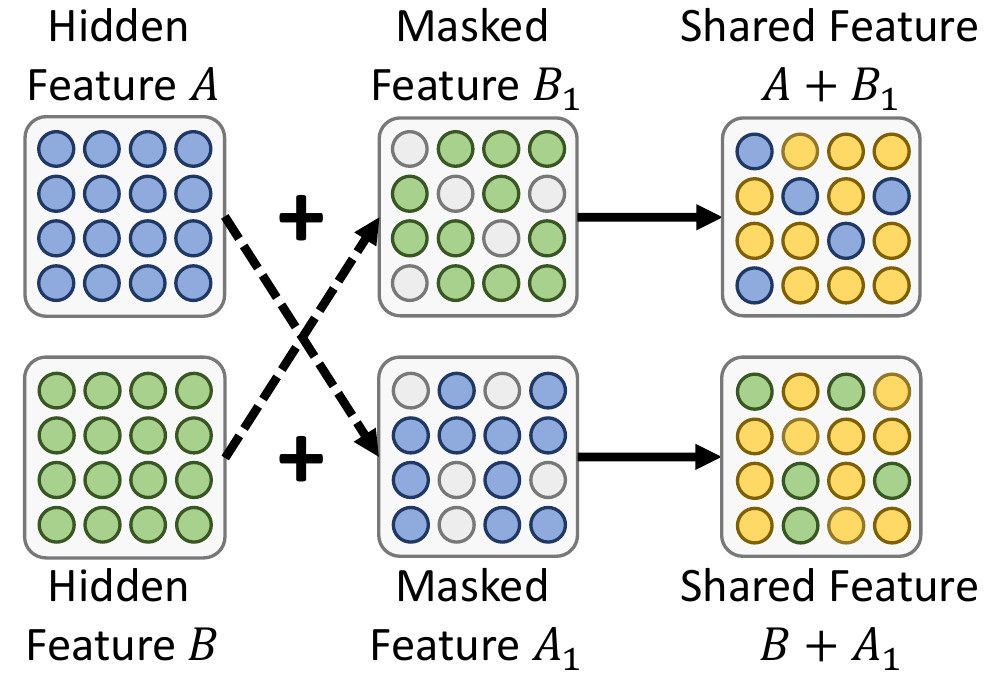, width=0.25\textwidth}\label{fig2:c}}
\caption{Histogram of the predictive entropy of model trained by (a) MCL (b) CMCL and (c) IE on CIFAR-10 and SVHN test data. 
{In the case of MCL and CMCL, we separate the classes of CIFAR-10 into specialized (i.e., classes that model has a class-wise test accuracy higher than 90$\%$) and non-specialized (others) classes.
In the case of IE, we follow the proposed method by \cite{lakshminarayanan2016simple}:
train an ensemble of 5 models with adversarial training (AT) and measure the entropy using the averaged probability, i.e., averaging output probabilities from 5 models.}
(d) Detailed view of feature sharing between two models. Grey units indicate that they are currently dropped. Masked features passed to a model are all added to generate the shared features.
}\label{fig2}
\end{figure*}

\subsection{Confident Oracle Loss}\label{sec:newloss}
In this section,
we propose a modified oracle loss for relaxing the issue of MCL described in the previous section. 
Suppose that the $m$-th model outputs the 
predictive distribution $P_{\theta_m}\left(y \mid \mathbf{x} \right)$
given input $\mathbf x$, where $\theta_m$ denotes
the model parameters. 
Then, we define the confident oracle loss as the following integer programming 
variant of \eqref{eq:oracleloss}:
\begin{subequations}
\label{eq:objective}
\begin{align}
L_{C}(\mathcal D)=
& ~\underset{v_i^m}{ \min} 
~ \sum \limits_{i=1}^N \sum \limits_{m=1}^M \Bigg(  v_i^m   \ell \left( y_i,P_{\theta_m}\left(y \mid \mathbf{x}_i \right) \right) \notag \\
&  +  \beta \left(1- v_i^m \right)
D_{{KL}} \left( {\mathcal U} \left( y \right) \parallel P_{\theta_{{ m}}}\left( y\mid \mathbf{x}_i \right) \right) 
\Bigg)   \label{eq:objective_fn} \\
 &\quad\text{subject to} \quad   \quad \sum_{m=1}^M v_i^m = 1, \quad \forall i, \label{eq:const} \\ 
&\quad\qquad\qquad\qquad v_i^m \in \{ 0,1 \}, \quad \forall i,m 
\end{align}
\end{subequations}
where 
$D_{KL}$ denotes the Kullback-Leibler (KL) divergence,
${\mathcal U} \left( y \right)$ is the uniform distribution, $\beta$ is a penalty parameter, 
and $v_i^m$ is a flag variable to decide the assignment of $\mathbf x_i$ to the $m$-th model.
By minimizing the KL divergence from the predictive distribution to the uniform one, 
the new loss forces the predictive distribution to
be closer to the uniform one, i.e., zero confidence, on non-specialized data,
while those for specialized data still follow the correct one.
For example, for classification tasks,
the most accurate model for each data is allowed to optimize the classification loss, 
while others are forced to give less confident predictions by minimizing the KL divergence.
We remark that although we optimize the KL divergence only for non-specialized data,
one can also do it even for specialized data to regularize each model
\cite{pereyra2017regularizing}.

\subsection{Stochastic Alternating Minimization for Training}\label{sec:alg}

In order to minimize the confident oracle loss \eqref{eq:objective} efficiently,
we use the following procedure \cite{guzman2012multiple},
which optimizes model parameters $\{ \theta_m \}$ and assignment variables $\{ v_i^m \}$ alternatively:
\begin{enumerate}
\item \noindent{\bf Fix $\{ \theta_m \}$ and optimize $\{ v_i^m \}$.}

Under fixed model parameters $\{ \theta_m \}$, the objective \eqref{eq:objective_fn}
is decomposable with respect to assignments $\{ v_i^m \}$ and 
it is easy to find optimal $\{ v_i^m \}$.

\item \noindent{\bf Fix $\{ v_i^m \}$ and optimize $\{ \theta_m \}$.}

Under fixed assignments $\{ v_i^m \}$,
the objective \eqref{eq:objective_fn}
is decomposable with respect to model parameters $\{ \theta_m \}$,
and it requires each model to be trained independently.
\end{enumerate}


The above scheme iteratively assigns each data to a particular model and then independently trains each model only using its assigned data.
Even though it monotonically decreases the objective, 
it is still highly inefficient
since it requires training each model multiple times until assignments $\{ v_i^m \}$ converge.
To address the issue, 
we propose deciding assignments and update model parameters to the gradient directions once per each batch, similarly to \citep{lee2016stochastic}. 
In other words, we perform a single gradient-update on parameters in Step 2, without waiting for their convergence to a (local) optimum. 
In fact, \citep{lee2016stochastic} show that such stochastic alternating minimization works well for the oracle loss \eqref{eq:oracleloss}.
We formally describe a detailed training procedure as the `version 0' of Algorithm \ref{alg:our},
and we will introduce the alternative `version 1' later.
This direction is complementary to ours, and we do not explore in this paper.


\begin{algorithm}[h!]
   \caption{Confident MCL (CMCL).}
   \label{alg:our}
\begin{algorithmic}
   \STATE {\bfseries Input:} Dataset $\mathcal{D}=\{ (\mathbf{x}_i, y_i) \mid \mathbf{x}_i \in \mathcal{X},y_i \in \mathcal{Y} \}$ and penalty parameter $\beta$
  \STATE {\bfseries Output:} Ensemble of $M$ trained models 
  \STATE \hrule
  \REPEAT
  \STATE Let ${\mathcal U} \left( y \right)$ be a uniform distribution
  \STATE Sample random batch $\mathcal{B} \subset \mathcal{D}$
  \FOR{$m=1$ {\bfseries to} $M$}
  \STATE Compute the loss of the $m$-th model:
  \STATE \hspace{0.1cm} \vspace{-0.1in}
  \begin{align*}
    L^m_i  \leftarrow & \beta \sum \limits_{{\widehat m} \neq m}  D_{KL} \left( {\mathcal U} \left( y \right) \parallel P_{\theta_{{ {\widehat m}}}}\left( y\mid \mathbf{x}_i \right) \right) \\
    &+ \ell \left( y_i,P_{\theta_m}\left(y_i \mid \mathbf{x}_i \right) \right), \quad \forall (\mathbf{x}_i, y_i) \in \mathcal{B}
  \end{align*}
  \ENDFOR
  \FOR{$m=1$ {\bfseries to} $M$}
  \FOR{$i=1$ {\bfseries to} $|\mathcal{B}|$}
  \IF{the $m$-th model has the lowest loss}
  \STATE Compute the gradient of the training loss $ \ell \left( y_i,P_{\theta_m}\left(y_i \mid \mathbf{x}_i \right) \right)$ w.r.t $\theta_m$
  \ELSE
  \STATE $/*$ version 0: exact gradient $*/$
  \STATE Compute the gradient of the KL divergence $\beta D_{KL} \left( {\mathcal U} \left( y \right) \parallel P_{\theta_{{ m}}}\left( y\mid \mathbf{x}_i \right) \right) $ w.r.t $\theta_m$
  \vspace{0.05in}
  \STATE $/*$ version 1: stochastic {labeling} $*/$
  \STATE Compute the gradient of the cross entropy loss $- \beta \log {P_{\theta_m}\left( {\widehat y}_i \mid \mathbf{x}_i \right)}$ using ${\widehat y}_i$ w.r.t $\theta_m$ where $\widehat y_i \sim {\mathcal U} \left( y \right)$
  \ENDIF
  \ENDFOR
  \STATE Update the model parameters
  \ENDFOR
   \UNTIL{$convergence$}
\end{algorithmic}
\end{algorithm}

\subsection{Effect of Confident Oracle Loss}\label{sec:justoracle}

Similar to Section \ref{ssec:oracleloss},
we evaluate the performance of the proposed training scheme 
using 5 CNNs for image classification on the CIFAR-10 dataset.
As shown in Figure \ref{fig1:c}, 
ensemble models trained by CMCL using the exact gradient (i.e., version 0 of Algorithm \ref{alg:our}) become specialists for certain classes. 
For specialized classes, 
they show the similar performance compared to the models trained by MCL, i.e., 
minimizing the oracle loss \eqref{eq:oracleloss},
which considers only specialization (see Figure \ref{fig1:b}).
For non-specialized classes,
ensemble
members of CMCL are not overconfident,
which makes it easy to pick a correct output via simple voting/averaging.
We indeed confirm that each model trained by CMCL has not only low entropy for its specialized classes, but also exhibits high entropy for non-specialized classes as shown in Figure \ref{fig2:b}.

We also evaluate the quality of confidence/uncertainty level on unseen data using SVHN \cite{11SVHN}.
Somewhat surprisingly, 
each model trained by CMCL only using CIFAR-10 training data exhibits high entropy for SVHN test data,
whereas models trained by MCL and IE are overconfident on it (see Figure \ref{fig2:a} and \ref{fig2:d}).
We emphasize that our method can produce confident predictions significantly
better than the proposed method by \citep{lakshminarayanan2016simple}, which uses the averaged probability of ensemble models trained by IE to obtain high quality uncertainty estimates (see Figure \ref{fig2:d}).

%% file: feature-sharing.tex
\section{Regularization Techniques}\label{sec:feature-gen}

In this section, we introduce advanced techniques 
for reducing the overconfidence and improving the performance.

\subsection{Feature Sharing}
\label{sec:feature-sharing}
We first propose a feature sharing scheme that stochastically shares the features among member models of CMCL to further
address the overconfidence issue.
The primary reason why deep learning models are overconfident is
that they do not always extract general features from data.
For examples, assume that some deep model only trains frogs and roses for classifying them.
Although there might exist many kinds of features on their images, 
the model might make a decision based only on some specific features, e.g., colors.
In this case, `red' apples can be classified as rose with high confidence.
Such an issue might be more severe in CMCL (and MCL) compared to IE since members of CMCL are specialized to certain data.
To address the issue, we suggest the feature ensemble approach that encourages each model to generate meaningful abstractions
from rich features extracted from other models.

Formally, consider an ensemble of $M$ neural networks with $L$ hidden layers.
We denote the weight matrix for layer $\ell$ of model $m\in [M]$ and $\ell$-th hidden feature of model $m$ by $\mathbf{W}^{\ell}_m$ and $\mathbf{h}^{\ell}_{m}$, respectively.
Instead of sharing the whole units of a hidden feature, we introduce random binary masks determining which units to be shared with other models.
We denote the mask for layer $\ell$ from model $n$ to $m$ as $\pmb{\sigma}_{nm}^{\ell} \sim \texttt{Bernoulli}(\lambda)$, which has the same dimension with $\mathbf{h}^{\ell}_{n}$ ({we use $\lambda=0.7$ in all experiments}).
Then, the $\ell$-th hidden feature of model $m$ with sharing $(\ell-1)$-th hidden features is defined as follows:
\begin{align*}
\mathbf{h}^{\ell}_{m} \left(\mathbf{x}\right) = \phi \left( \mathbf{W}_m^{\ell} \left( \mathbf{h}^{\ell-1}_{m} \left(\mathbf{x}\right) + \sum \limits_{n \neq m} \pmb{\sigma}_{nm}^{\ell} \star \mathbf{h}^{\ell-1}_{n} \left(\mathbf{x}\right) \right) \right),
\end{align*}
where $\star$ denotes element-wise multiplication and $\phi$ is the activation function.
Figure \ref{fig2:c} illustrates the proposed feature sharing scheme in an ensemble of deep neural networks.
It makes each model learn more generalized features by sharing the features among them.
However, one might expect that it might make each model overfitted due to the increased number of  parameters that induces a single prediction, i.e., the statistical dependencies among outputs of models increase, which would hurt the ensemble effect.
In order to handle this issue, we introduce the randomness in sharing across models in a similar manner to DropOut \cite{srivastava2014dropout} using the random binary masks $\pmb{\sigma}$.
In addition, we propose sharing features at lower layers since sharing the higher layers might overfit the overall networks more.
For example, in all experiments with CNNs in this paper, we commonly apply feature sharing for hidden features just before the first pooling layer.
We also remark that such feature sharing strategies for better generalization have also been investigated in the literature for different purposes \cite{misra2016cross,rusu2016progressive}.

\subsection{Stochastic Labeling}

For more efficiency in minimizing the confident oracle loss,
we also consider a noisy unbiased estimator of gradients of the KL divergence with Monte Carlo samples from the uniform distribution.
The KL divergence from the predictive distribution to the uniform distribution can be written as follows:
\begin{align*}
& D_{KL} \left( \mathcal{U} \left( y \right) \parallel P_{\theta}\left( y\mid \mathbf{x} \right) \right) \\
& = \sum \limits_y  \mathcal{U} \left( y \right) \log \frac{ \mathcal{U} \left( y \right)}{P_{\theta}\left( y\mid \mathbf{x} \right)} \\
&= \sum \limits_y  \mathcal{U} \left( y \right) \log  \mathcal{U} \left( y \right) - \sum \limits_y  \mathcal{U} \left( y \right) {\log P_{\theta}\left( y\mid \mathbf{x} \right)}.
\end{align*}
Hence, the gradient of the above KL divergence
with respect to the model parameter $\theta$ becomes 
\begin{align*}
 \bigtriangledown_{\theta} D_{KL} \left(  \mathcal{U} \left( y \right) \parallel P_{\theta}\left( y\mid \mathbf{x} \right) \right) 
& = - \E_{ \mathcal{U} \left( y \right)} [ \bigtriangledown_{\theta} {\log P_{\theta}\left( y\mid \mathbf{x} \right)} ].
\end{align*}
From the above, we induce the following noisy unbiased estimator of gradients with Monte Carlo samples from the uniform distribution:
\begin{align*}
& - \E_{ \mathcal{U} \left( y \right)} [ \bigtriangledown_{\theta} {\log P_{\theta}\left( y\mid \mathbf{x} \right)} ]
\backsimeq - \frac{1}{S} \sum_s \bigtriangledown_{\theta} {\log P_{\theta}\left( y^{s}\mid \mathbf{x} \right)},
\end{align*}
where $y^{s} \sim  \mathcal{U} \left( y \right)$ and $S$ is the number of samples.
This random estimator takes samples from the uniform distribution $\mathcal{U} \left( y \right)$ 
and constructs estimates of the gradient using them.
In other words, $\bigtriangledown_{\theta} {\log P_{\theta}\left( y^{s}\mid \mathbf{x} \right)}$
is the gradient of the cross entropy loss under assigning a random label 
to $\mathbf x$.
This stochastic labeling provides efficiency in implementation/computation
and 
stochastic regularization effects.
We formally describe detailed procedures, 
as the ‘version 1’ of Algorithm \ref{alg:our}.

%% file: experiments.tex
\section{Experiments}
\label{sec:exp}

We evaluate our algorithm for both classification and foreground-background segmentation tasks using CIFAR-10 \cite{cifar09}, SVHN \cite{11SVHN} and iCoseg \cite{icoseg10} datasets.
In all experiments, we compare the performance of CMCL with those of traditional IE and MCL using deep models. 
We provide the more detailed experimental setups including model architectures in the supplementary material.\footnote{Our code is available at \url{https://github.com/chhwang/cmcl}.}

\subsection{Image Classification} \label{ssec:exp-cls}
\noindent {\bf Setup.}
The CIFAR-10 dataset consists of 50,000 training and 10,000 test images with 10 image classes where each image consists of $32\times32$ RGB pixels.
The SVHN dataset consists of 73,257 training
and 26,032 test images.\footnote{We do not use the extra SVHN dataset for training.}
We pre-process the images with global contrast normalization and ZCA whitening 
following \cite{maxout,wrn16}, and do not use any data augmentation.
Using these datasets, we train various CNNs, e.g., VGGNet \cite{vggnet}, GoogLeNet \cite{googlenet}, and ResNet \cite{resnet}.
Similar to \citep{wrn16}, we use the softmax classifier, and train each model by minimizing the cross-entropy loss using the stochastic gradient descent method with Nesterov momentum.

For evaluation, we measure the top-1 and oracle error rates on the test dataset.
The top-1 error rate is calculated by averaging output probabilities from all models and predicting the class of the highest probability.
The oracle error rate is the rate of classification failure over all outputs of individual ensemble members for a given input, i.e., it measures whether none of the
members predict the correct class for an input.
While a lower oracle error rate suggests higher diversity, a lower oracle error rate does not always bring a higher top-1 accuracy as this metric does not reveal the level of overconfidence of each model.
By collectively measuring the top-1 and oracle error rates, one can grasp the level of specialization and confidence of a model.

\begin{table}[t]
\centering
\resizebox{\columnwidth}{!}{
\begin{tabular}{ccccc}
\hline
\begin{tabular}[c]{@{}c@{}}Ensemble\\Method\end{tabular}&\begin{tabular}[c]{@{}c@{}}Feature\\ Sharing\end{tabular}
&\begin{tabular}[c]{@{}c@{}}Stochastic\\ Labeling\end{tabular} &
\begin{tabular}[c]{@{}c@{}}Oracle\\ Error Rate\end{tabular}&
\begin{tabular}[c]{@{}c@{}}Top-1\\ Error Rate\end{tabular} \\ \hline
IE    & -            & -           & 10.65\%       & 15.34\%       \\ \hline
MCL   & -            & -           & 4.40\%        & 60.40\%       \\ \hline
\multirow{3}{*}{CMCL}
            & -              & -             & 4.49\%        & 15.65\%       \\
            & \checkmark     & -             & 5.12\%        & 14.83\%       \\
            & \checkmark     & \checkmark    & \bf{3.32\%}   & \bf{14.78\%}  \\ \hline
\end{tabular}}
\caption{Classification test set error rates on CIFAR-10 using various ensemble methods.
}
\label{tbl:comparison}
\end{table}

\noindent {\bf Contribution by each technique.} 
Table \ref{tbl:comparison} validates contributions of our suggested techniques under comparison 
with other ensemble methods IE and MCL. We evaluate an ensemble of five simple CNN models where each model has two convolutional layers followed by a fully-connected layer. 
We incrementally apply our optimizations to gauge the stepwise improvement by each component. One can note that CMCL significantly outperforms MCL in the top-1 error rate even without feature sharing or stochastic labeling while it still provides a comparable oracle error rate.
By sharing the 1st ReLU activated features, 
the top-1 error rates are improved compared to those that employ only confident oracle loss.
Stochastic labeling 
further improves both error rates. 
This implies that stochastic labeling not only reduces computational burdens but also provides  regularization effects. 

\begin{table*}[t]
\centering
\begin{tabular}{ccllll}
\hline
\multirow{2}{*}{Ensemble Method}   & \multirow{2}{*}{$K$} & \multicolumn{2}{l}{\hspace{12mm}Ensemble Size $M=5$}  & \multicolumn{2}{l}{\hspace{11mm}Ensemble Size $M=10$}     \\
                                   &                    & Oracle Error Rate & Top-1 Error Rate    & Oracle Error Rate & Top-1 Error Rate    \\ \hline
IE  & -  & \hspace{7mm}10.65\%                   & \hspace{6mm}15.34\%                    & \hspace{6mm}9.26\%                  & \hspace{6mm}15.34\%           \\ \hline
\multirow{4}{*}{MCL} 
             & 1  & \hspace{8mm}4.40\%                    & \hspace{6mm}60.40\%                    & \hspace{6mm}\textbf{0.00\%}    & \hspace{6mm}76.88\%            \\
             & 2 & \hspace{8mm}3.75\%                    & \hspace{6mm}20.66\%                    & \hspace{6mm}1.46\%                   & \hspace{6mm}49.31\%                    \\
             & 3 & \hspace{8mm}4.73\%                    & \hspace{6mm}16.24\%                    & \hspace{6mm}1.52\%                 & \hspace{6mm}22.63\%                    \\
             & 4 & \hspace{8mm}5.83\%                    & \hspace{6mm}15.65\%                    & \hspace{6mm}1.82\%                   & \hspace{6mm}17.61\%                    \\ \hline
\multirow{4}{*}{CMCL}
            &  1 & \hspace{8mm}\textbf{3.32\%}          & \hspace{6mm}14.78\%           & \hspace{6mm}1.96\%                   & \hspace{6mm}14.28\%                    \\
            &  2 & \hspace{8mm}3.69\% & \hspace{6mm}\textbf{14.25\% (-7.11\%)}          & \hspace{6mm}1.22\%                   & \hspace{6mm}13.95\%                   \\
            &  3 & \hspace{8mm}4.38\%                    & \hspace{6mm}14.38\%           & \hspace{6mm}1.53\%                   & \hspace{6mm}14.00\%                    \\
            &  4 & \hspace{8mm}5.82\%                    & \hspace{6mm}14.49\%           & \hspace{6mm}1.73\%                  & \hspace{6mm}\textbf{13.94\% (-9.13\%)} \\ \hline
\end{tabular}
\caption{
Classification test set error rates on CIFAR-10 with varying values of the overlap parameter $K$ explained in Section \ref{ssec:exp-cls}. 
We use CMCL with both feature sharing and stochastic labeling.
Boldface values in parentheses represent the relative reductions from the best results of MCL and IE.
}\label{tbl:cifar10}
\end{table*}

\begin{table*}[]
\centering
\begin{tabular}{cccllll}
\hline
\multirow{2}{*}{Model Name}               &
\multirow{2}{*}{\begin{tabular}[c]{@{}c@{}}Ensemble\\Method\end{tabular}} &
\multicolumn{2}{l}{\hspace{18mm}CIFAR-10} &
\multicolumn{2}{l}{\hspace{22mm}SVHN} \\
& &
Oracle Error Rate  &
Top-1 Error Rate &
Oracle Error Rate  &
Top-1 Error Rate   \\ \hline
\multirow{4}{*}{VGGNet-17}    & - (single)           & \hspace{0mm}10.65\%                    &\hspace{6mm}10.65\%    & \hspace{8mm}5.22\%                    & \hspace{7mm}5.22\%       \\
& IE           & \hspace{0mm}3.27\%                    & \hspace{7mm}8.21\%    & \hspace{8mm}1.99\%                    & \hspace{7mm}4.10\%       \\
                           & MCL          & \hspace{0mm}\textbf{2.52\%}                    & \hspace{6mm}45.58\%    & \hspace{8mm}\textbf{1.45\%}                    & \hspace{6mm}45.30\%         \\
                           & CMCL         & \hspace{0mm}2.95\% & \hspace{7mm}\textbf{7.83\% (-4.63\%)}  & \hspace{8mm}1.65\% & \hspace{7mm}\textbf{3.92\% (-4.39\%)} \\ \hline
\multirow{4}{*}{GoogLeNet-18} & - (single)           & \hspace{0mm}10.15\%                    & \hspace{6mm}10.15\%    & \hspace{8mm}4.59\%                    & \hspace{7mm}4.59\%       \\
& IE           & \hspace{0mm}3.37\%                    & \hspace{7mm}7.97\%    & \hspace{8mm}1.78\%                    & \hspace{7mm}3.60\%       \\
                           & MCL          & \hspace{0mm}\textbf{2.41\%}           & \hspace{6mm}52.03\% & \hspace{8mm}1.39\%           & \hspace{6mm}37.92\%         \\
                           & CMCL         & \hspace{0mm}2.78\%                    & \hspace{7mm}\textbf{7.51\% (-5.77\%)}
                           &\hspace{8mm}\textbf{1.36\%}                    & \hspace{7mm}\textbf{3.44\% (-4.44\%)}  \\ \hline
\multirow{4}{*}{ResNet-20} & - (single)           & 14.03\%                    & \hspace{6mm}14.03\%    & \hspace{8mm}5.31\%                    & \hspace{7mm}5.31\%       \\
& IE           & \hspace{0mm}3.83\%                    & \hspace{6mm}10.18\% & \hspace{8mm}1.82\%                    & \hspace{7mm}3.94\%         \\
                           & MCL          & \hspace{0mm}\textbf{2.47\%}           & \hspace{6mm}53.37\% & \hspace{8mm}1.29\%           & \hspace{6mm}40.91\%         \\
                           & CMCL         & \hspace{0mm}2.79\%                    & \hspace{7mm}\textbf{8.75\% (-14.05\%)}
                           & \hspace{8mm}\textbf{1.42\%}                    & \hspace{7mm}\textbf{3.68\% (-6.60\%)}\\ \hline
\end{tabular}
\caption{Classification test set error rates on CIFAR-10 and SVHN for various large-scale CNN models. We train an ensemble of 5 models, and use CMCL with both feature sharing and stochastic labeling.
Boldface values in parentheses indicate relative error rate reductions from the best results of MCL and IE.}
\label{tbl:deep}
\end{table*}

\noindent {\bf Overlapping.}
As a natural extension of CMCL,
we also consider picking $K$ specialized models instead of having only one specialized model, 
which was investigated for original MCL~\citep{guzman2012multiple,lee2016stochastic}. This is easily achieved by modifying the constraint \eqref{eq:const} as $\sum_{m=1}^M v_i^m = K$, where $K$ is an overlap parameter that controls training data overlap between the models. This simple but natural scheme brings extra gain in top-1 performance by generalizing each model better. 
Table~\ref{tbl:cifar10} compares the performance of various ensemble methods with varying values of $K$.
Under the choice of $K=4$, CMCL of 10 CNNs provides 9.13\% relative reduction in the top-1 error rates from the corresponding IE.
Somewhat interestingly, 
IE has similar error rates on ensembles of 
both 5 and 10 CNNs, which implies that the performance of CMCL might be impossible to achieve
using IE even if one increases the number of models in IE.

\begin{figure*}[t]
\includegraphics[width=\textwidth]{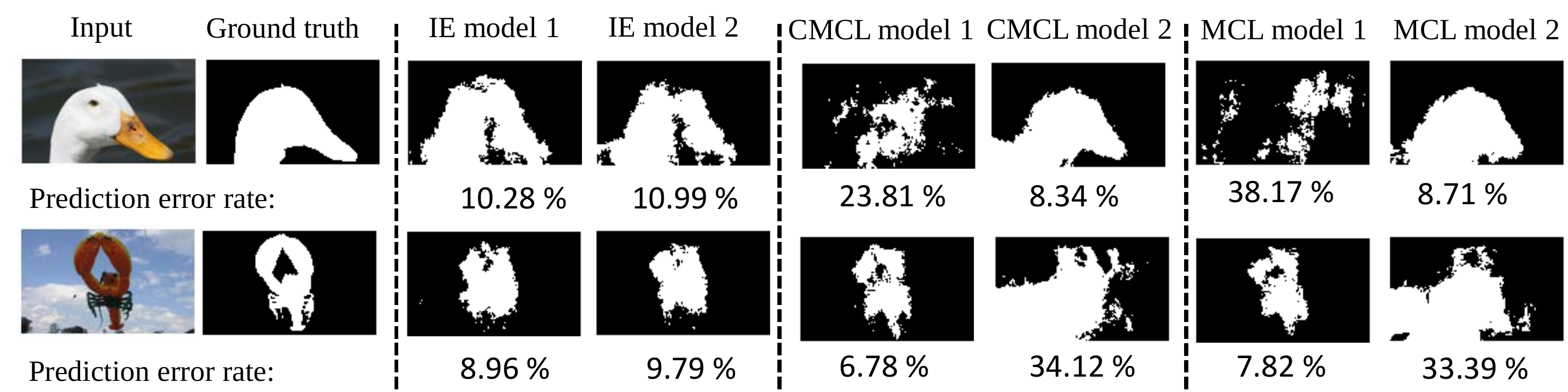}
\caption{Prediction results of foreground-background segmentation for a few sample images. 
A test error rate is shown below each prediction.
The ensemble models trained by CMCL and MCL generate high-quality predictions specialized for certain images.
%
}\label{fig:icoseg}
\end{figure*}

\begin{figure*} [t] \centering
\subfigure[]
{\epsfig{file=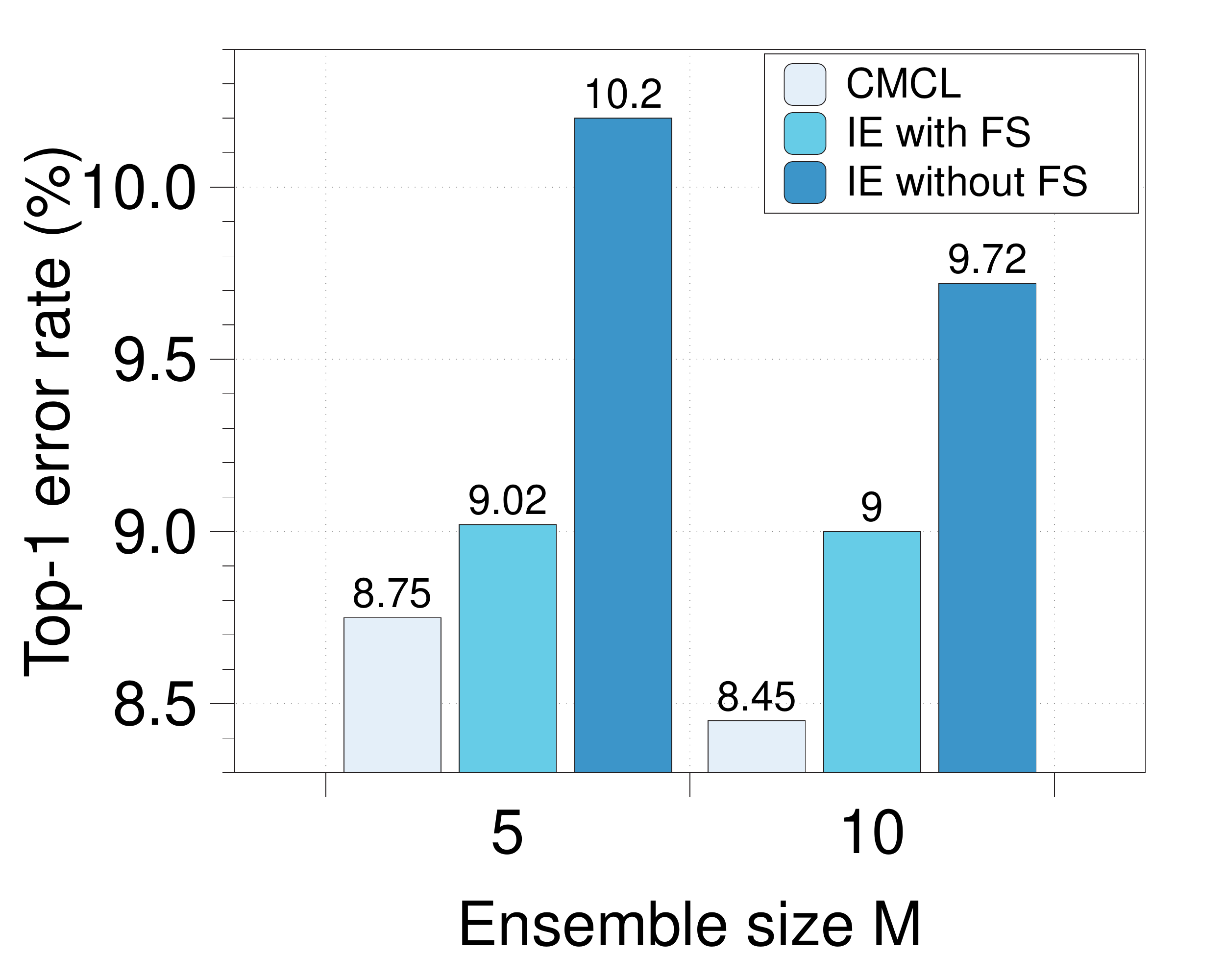, width=0.32\textwidth}\label{fig5:c}}
\,
\subfigure[]
{\epsfig{file=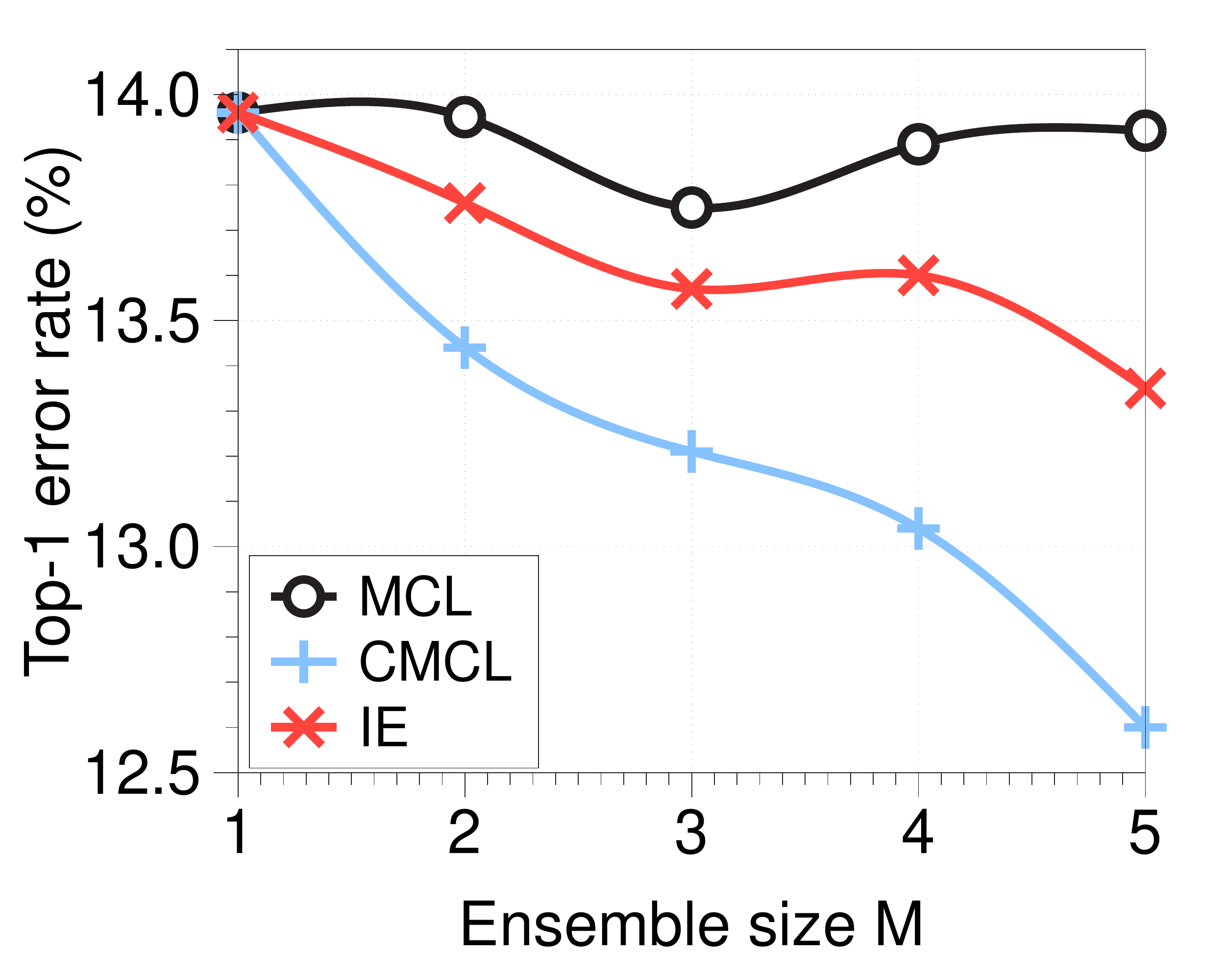, width=0.32\textwidth}\label{fig5:a}}
\,
\subfigure[]
{\epsfig{file=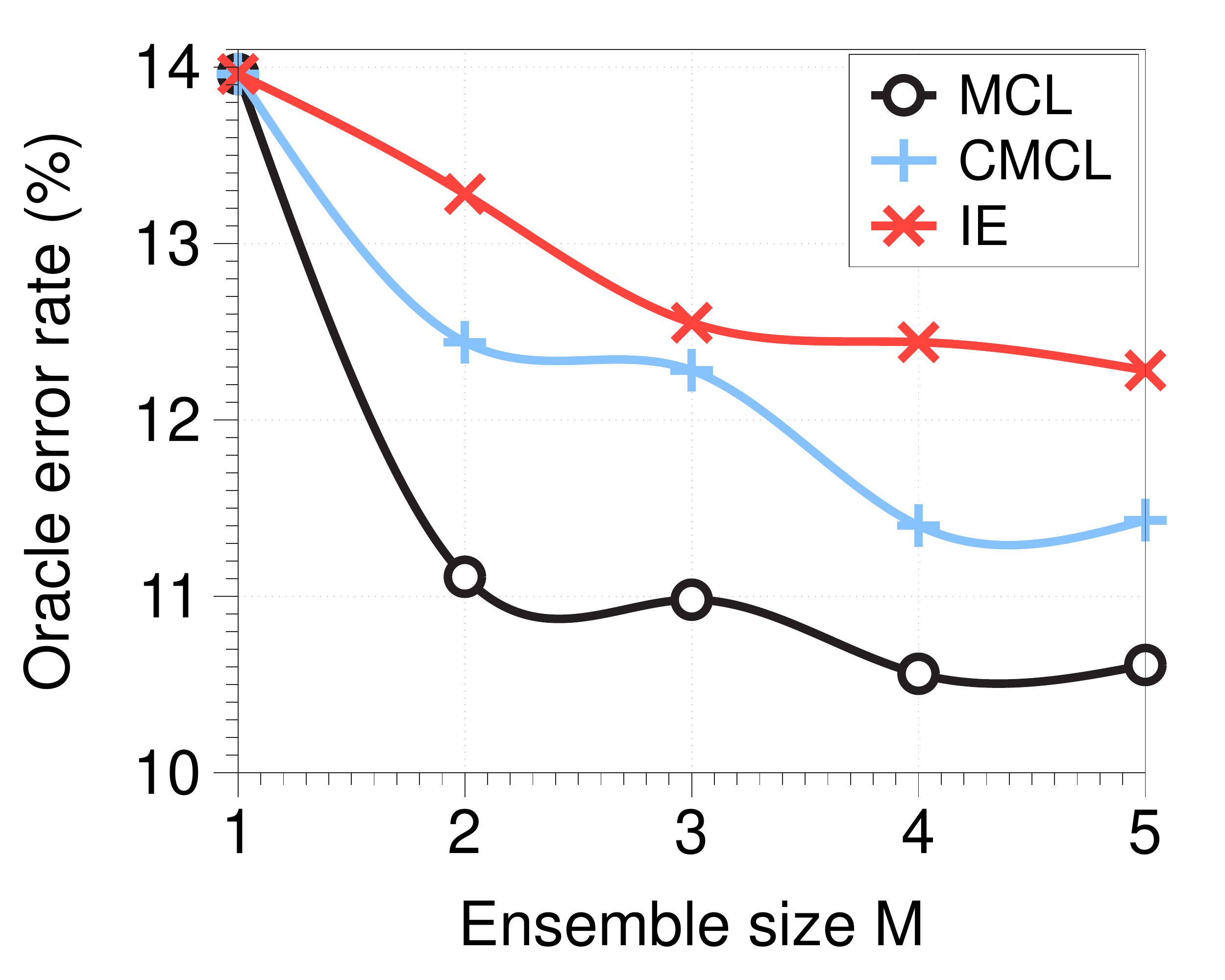, width=0.32\textwidth}\label{fig5:b}}
\caption{(a) Top-1 error rate on CIFAR-10. We train an ensemble of $M$ ResNets with 20 layers, and apply feature sharing (FS) to IE and CMCL. (b) Top-1 error rate and (c) oracle error rate on iCoseg
by varying the ensemble sizes. 
The ensemble models trained by CMCL consistently improves the top-1 error rate over baselines.
}\label{fig:icosef}
\end{figure*}
\noindent {\bf Large-scale CNNs.} 
We now evaluate the performance of our ensemble method when it is applied to larger-scale CNN models for image classification tasks on CIFAR-10 and SVHN datasets.
Specifically, we test VGGNet~\cite{vggnet}, GoogLeNet~\cite{googlenet}, and ResNet~\cite{resnet}. 
We share the non-linear activated features right before the first pooling layer,
i.e., the 6th, 2nd, and 1st ReLU activations for ResNet with 20 layers, VGGNet with 17 layers, and GoogLeNet with 18 layers, respectively.
This choice is for maximizing the regularization effect of feature sharing while minimizing the statistical dependencies among the ensemble models. 
For all models, we choose the best hyper-parameters for confident oracle loss among the penalty parameter $\beta \in \{0.5,0.75,1,1.25,1.5\}$ and the overlapping parameter $K\in\{2,3,4\}$. 
Table~\ref{tbl:deep} shows that CMCL consistently outperforms all baselines with respect to the top-1 error rate while producing comparable oracle error rates to those of MCL.
We also apply the feature sharing to IE as reported in Figure \ref{fig5:c}.
Even though the feature sharing also improves the performance of IE, CMCL still outperforms IE:
CMCL provides 6.11\% relative reduction of the top-1 error rate from the IE with feature sharing under the choice of $M=10$. We also remark that IE with feature sharing has similar error rates as the ensemble size increases, while CMCL does not (i.e., the gain is more significant for CMCL).
This implies that feature sharing is more effectively working for CMCL.

\subsection{Foreground-Background Segmentation}\label{ssec:exp-icoseg}

In this section, we evaluate if ensemble models trained with CMCL produce high-quality segmentation of foreground and background of an image with the iCoseg dataset.
The foreground-background segmentation is formulated as a pixel-level classification problem with 2 classes, i.e., 0 (background) or 1 (foreground). To tackle the problem, we design fully convolutional networks (FCNs) model \cite{long2015fully} based on the decoder architecture presented in \cite{radford2015unsupervised}.
The dataset consists of 38 groups of related images with pixel-level ground truth on foreground-background segmentation of each image. We only use images that are larger than $300 \times 500$ pixels.
For each class, we randomly split $80\%$ and $20\%$ of the data into training and test sets, respectively. We train on $75 \times 125$ resized images using the bicubic interpolation \cite{keys1981cubic}. 
Similar to \cite{guzman2012multiple,lee2016stochastic}, we initialize the parameters of FCNs with those trained by IE for MCL and CMCL.
For all experiments, CMCL is used with both feature sharing and stochastic labeling.

Similar to \cite{guzman2012multiple}, we define the percentage of incorrectly labeled pixels as prediction error rate.
We measure the oracle error rate (i.e., the lowest error rate over all models for a given input) and the top-1 error rate.
The top-1 error rate is measured by following the predictions of the member model that has a lower pixel-wise entropy, i.e., picking the output of a more confident model.
For each ensemble method, we vary the number of ensemble models and measure the oracle error rate and test error rate.
Figure~\ref{fig5:a} and \ref{fig5:b} show both top-1 and oracle error rates for all ensemble methods. We remark that the ensemble models trained by CMCL consistently improves the top-1 error rate over baselines. In an ensemble of 5 models, we find that CMCL achieve up to $6.77\%$ relative reduction in the top-1 error rate from the corresponding IE.
As shown in Figure \ref{fig:icoseg}, an individual model trained by CMCL generates high-quality solutions by specializing itself in specific images (e.g., model 1 is specialized for `lobster' while model 2 is specialized for `duck') while each model trained by IE does not.

%% file: conclusion.tex
\section{Conclusion}
\label{sec:conclusion}

This paper proposes CMCL, a novel ensemble method of deep neural networks that produces diverse/plausible confident prediction of high quality. 
To this end, we address the over-confidence issues of MCL,
and propose a new loss, architecture and training method.
In our experiments, CMCL outperforms not only the known MCL, but also
the traditional IE, with respect to the top-1 error rates
in classification and segmentation tasks.
The recent trend in the deep learning community tends to 
make models bigger and wider. We believe that our new ensemble approach brings a refreshing angle for developing advanced large-scale deep neural networks in many related applications.

%% file: Ack.tex
\section*{Acknowledgements} 
 This work was supported in part by the ICT R\&D Program of MSIP/IITP, Korea, under [2016-0-00563, Research on Adaptive Machine Learning Technology Development for Intelligent Autonomous Digital Companion], R0190-16-2012, [High Performance Big Data Analytics Platform Performance Acceleration Technologies Development], and by the National Research Council of Science \& Technology (NST) grant by the Korea government (MSIP) (No. CRC-15-05-ETRI).

%% file: appendix.tex
\appendix
\onecolumn
\clearpage
\begin{center}{\bf {\LARGE Supplementary Material:}}
\end{center}

\begin{center}{\bf {\Large Confident Multiple Choice Learning}}
\end{center}

\section{Experimental Setups for Image Classification}
In this section, we describe detailed explanation about all the experiments described in Section \ref{ssec:exp-cls}.

\noindent {\bf Detailed CNN structure and training.}

The CNN we use for evaluations in Table \ref{tbl:comparison} is consist of two convolutional layers followed by one fully-connected layer.
Convolutional layers have 128 and 256 filters respectively.
Each convolutional layer has a $5\times5$ receptive field applied with a stride of 1 pixel.
Each max pooling layer pools $2\times2$ regions at strides of 2 pixels.
Dropout was applied to all layers in the network with drop probability $0.5$.
Similar to \citep{wrn16}, the softmax classifier is used, and each model is trained by minimizing the cross-entropy loss using SGD with Nesterov momentum. 
The initial learning rate is set to 0.01, weight decay to 0.0005, dampening to 0, momentum to 0.9 and minibatch size to 64.
We drop the learning rate by 0.2 at 60, 120 and 160 epochs and we train for total 200 epochs.
We report the mean of the test error rates produced by repeating each test 5 times.

\begin{figure*} [h] \centering
\subfigure[VGGNet-17 overview]
{\epsfig{file=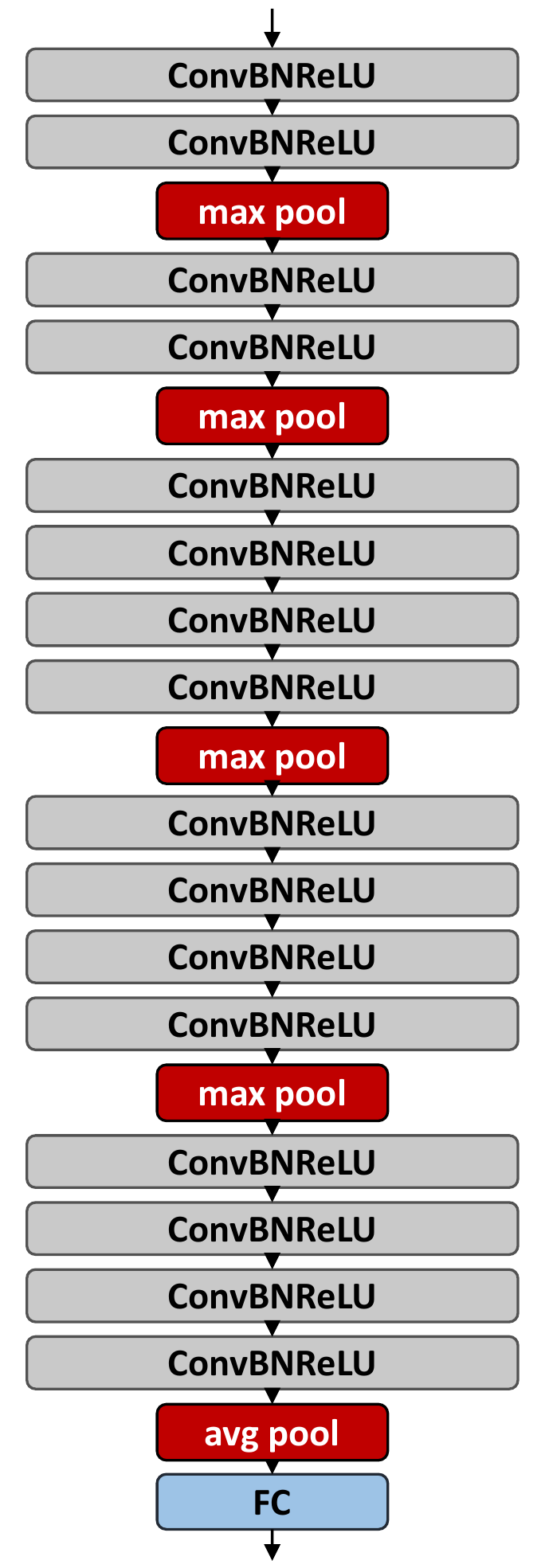, width=0.2\textwidth}\label{fig:vggnet}}
\,
\subfigure[GoogLeNet-18 overview]
{\epsfig{file=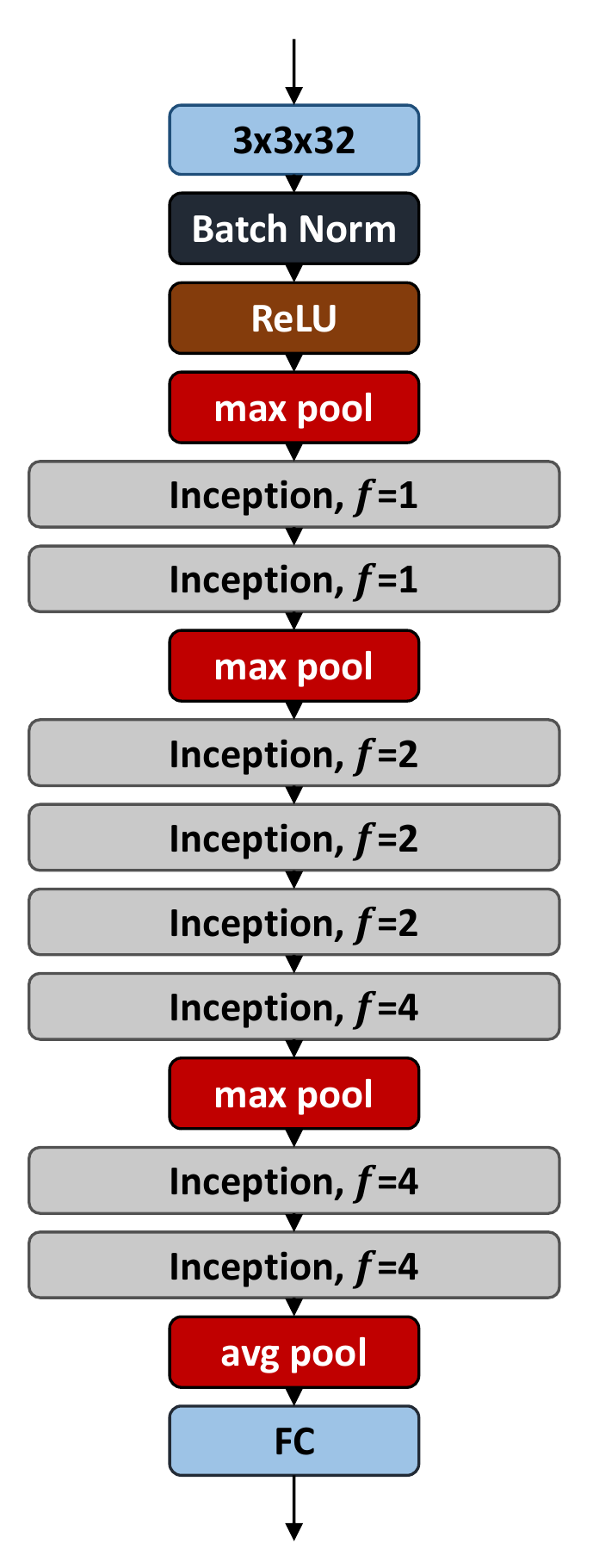, width=0.22\textwidth}\label{fig:googlenet}}
\,
\subfigure[Structure of an inception module with width factor $f$]
{\epsfig{file=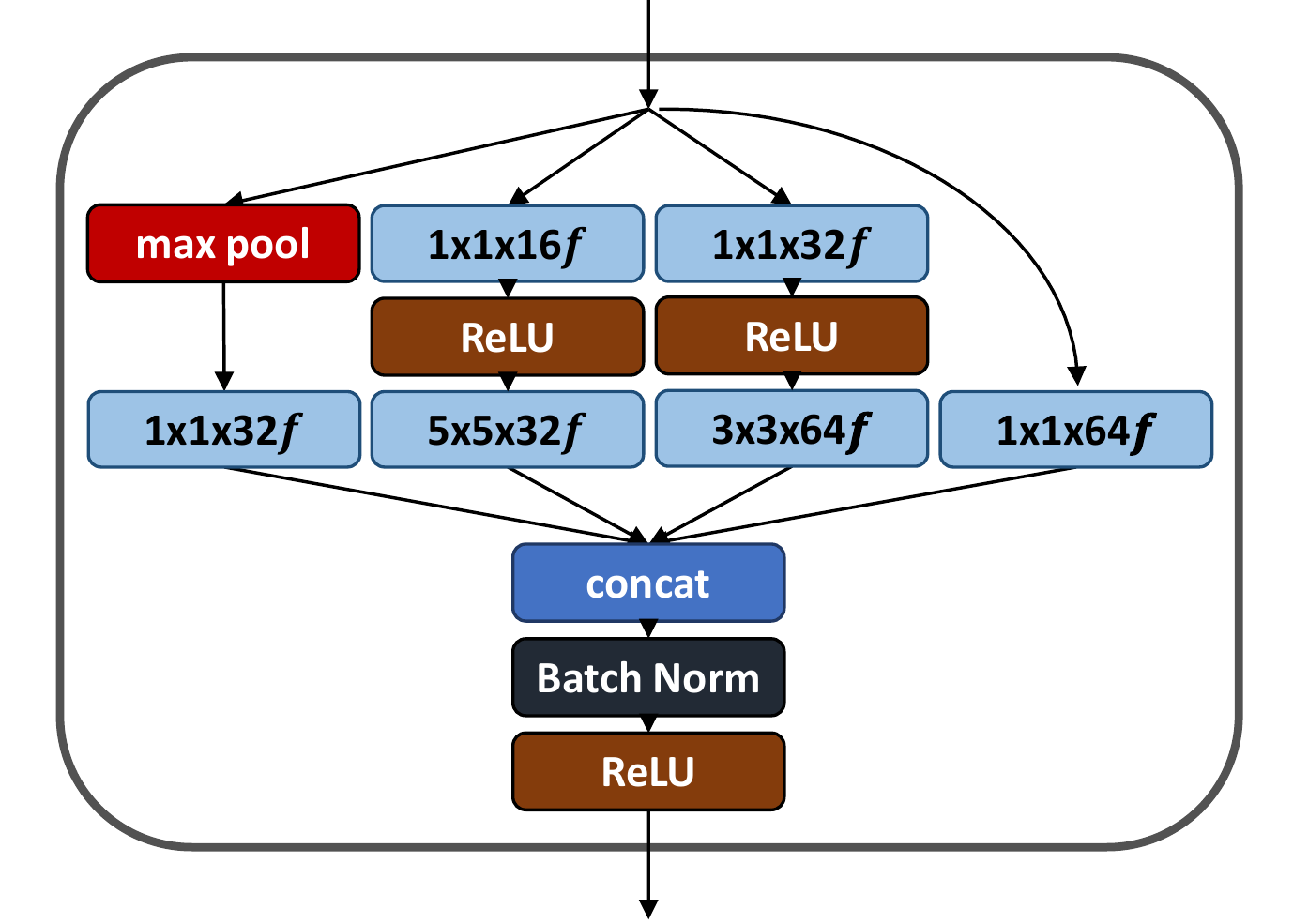, width=0.5\textwidth}\label{fig:inception}}
\caption{Detailed structure of large-scale CNNs used in Section \ref{ssec:exp-cls}.}\label{fig:deepnets}
\end{figure*}

\noindent {\bf Detailed large-scale CNN models.}

In case of residual networks, we use ResNet-20 model suggested by the author, which has 19 $3\times3$ convolutional layers.
We also follow the author's descriptions to train the model: minibatch size is set to 128, weight decay to 0.0001, momentum to 0.9, and initial learning rate to 0.1 and drop by 0.1 after 82 and 123 epochs with 164 epochs in total.
Figure \ref{fig:vggnet} shows the detailed structure of VGGNet-17 with one fully-connected layer and 16 convolutional layers.
Each ConvBNReLU box in the figure indicates a $3\times3$ convolutional layer followed by batch normalization \cite{batchnorm} and ReLU activation.
Figure \ref{fig:googlenet} shows the detailed structure of GoogLeNet-18 with one fully-connected layer and 8 inception modules consist of 17 convolutional layers in total, where $1\times1$ convolutional layers are not considered as weighted layers.
To simply increase the number of convolutional filters as layers stacked on, we introduce width factor $f$ which controls the overall size of an inception module as shown in Figure \ref{fig:inception}.
For both VGGNet-17 and GoogLeNet-18, all convolutional layers have stride 1 and use padding to keep the feature map size equal.
Also, all max pooling layers have $3\times3$ receptive fields with stride 1 and all average pooling layers indicate the global average pooling \cite{nin}.
We use initial learning rate 0.1 and drop it by 0.2 at 25, 50 and 75 epochs with total 100 epochs for both networks.
We use Nesterov momentum 0.9 for SGD, minibatch size is set to 128, and weight decay is set to 0.0005.
We report the mean of the test error rates produced by repeating each test 5 times.

\section{Experimental Setups for Background-Foreground Segmentation}
In this section, we describe detailed explanation about all the experiments described in Section \ref{ssec:exp-icoseg}.
It consists of three convolutional layers followed by a fully convolutional layer.  
The convolutional layers have 128, 256 and 1 filters respectively. Each convolutional layer has a $4 \times 4$ receptive field applied with a stride of 2 pixel. 
For feature sharing, the 2-th activation of FCNs is used.
The softmax classifier is used, and each model is trained by minimizing the cross-entropy loss using Adam learning rule \cite{kingma2014adam} with a mini-batch size of 20. The initial learning rate is chosen from $\{0.001,0.0005,0.0001\}$ and we used an exponentially decaying learning rate.
We train every model for total 300 epochs.
Similar to \cite{guzman2012multiple,lee2016stochastic},
we initialize the parameter of FCNs using that of FCNs trained by IE for 20 epochs in case of MCL and CMCL.
The best test result is reported for each method.

